\documentclass{ceurart}

\ifdefined\pdfsuppressptexinfo \pdfsuppressptexinfo=-1 \fi

\usepackage{url}

\begin{document}

\copyrightyear{2026}
\copyrightclause{Copyright for this paper by its authors.
  Use permitted under Creative Commons License Attribution 4.0
  International (CC BY 4.0).}

\conference{LLM4XAI 2026: Workshop on Large Language Models for Explainable AI,
  co-located with CIKM 2026, November 8, 2026}

\title{Does a model's stated reason for rejecting a candidate do any work?}

\author[1]{Archit Rastogi}[%
email=architrastogi20@gmail.com,
]
\address[1]{Independent Researcher}

\begin{abstract}
Asked to choose between candidates and explain the choice, a language model often rejects a rival
by naming a fact its profile lacks: no director, no date of death. That sentence is a claim about
the text in front of the model, and it can be tested without any judge. We insert a real corpus
sentence stating the named fact into the rival's profile and ask again under greedy decoding. Two
controls separate content from placement: a length-matched irrelevant sentence at the same
profile, and the same two sentences at a third option the model never mentioned. In the largest of three
runs, six open models on 2WikiMultihopQA, supplying the named fact at the profile the model
named moves its choice more than the irrelevant control does, odds ratio 3.57 [1.54, 8.26], Holm
$p=0.0210$%
, and this survives dropping any single model. The contrast the design was built to detect, the
same fact at the option nobody named, does not clear correction, Holm $p=0.2428$%
. The strongest result in the family carries no content claim at all: the identical irrelevant
sentence moves the choice more at the named rival than at the third option, Holm $p=0.0008$%
. Repair and control also differ in co-candidate mentions, relation template and fluency; post-hoc matching on the first two
preserves the content effects' direction, matching fluency weakens one, so the content contrasts bound
an effect rather than establish one. A forced single-token probability read disagrees in
direction with the free-text choice on that same contrast, and three candidate explanations for
the disagreement find no support. Every measurement is a string rule, so each was validated
against the records it reads; validation caught eight defects. The largest, a choice-parsing
rule that returned the option a model had just rejected in 17.1\% of adjudicable responses%
, would have reported six surviving contrasts instead of four. Code, per-item records with every
raw response, and seeded analysis scripts are released.
\end{abstract}


\maketitle

\section{Introduction}
\label{sec:intro}

When a person explains a decision by ruling an option out, they usually name something it lacks.
That sentence carries its own test: supply the missing thing and see whether the decision
changes. This is a contrastive explanation, studied in explainable AI before large language
models existed \citep{miller2019explanation}. Large language models, asked to pick a candidate
and explain the pick, produce the same form: asked why a rival was rejected, they name a
concrete fact its profile lacks. Because the named fact is a claim about text the model can see,
the test needs no judge. Insert the fact, ask again, and read the choice.

A real item from this study makes the setting concrete (Figure~\ref{fig:design}). It comes from
the six-option design: the question is ``Which film has the director died later, The Hole (1960
Film) or Nootrukku Nooru?'', and the model sees six candidates with short profiles: A) The Hole
in the Wall (1921 film), B) Thulasi (1987 film), C) The Subterraneans (film), D) Le Masque de la
M\'eduse, E) Nootrukku Nooru, F) Querelle. Llama-3.1-8B-Instruct chooses E and, asked what it
ruled out, says ``I ruled out option A, The Hole in the Wall, because the profile does not
mention the director's death.'' The named defect is a missing date of death. The repair (R1)
appends to A's profile a sentence stating one, taken from sibling D's profile with the subject
swapped: ``The Hole in the Wall (1921 film) was Rollin's final film, as the director died in
2010.'' The sentence is not true of A, and need not be: the claim under test is only that the
profile omits the attribute (Section~\ref{sec:method}). The control (R2) appends an unrelated
corpus sentence of about the same length to A instead. The same two sentences are appended to C,
an option the model never mentioned (R3, R4), and an unedited re-ask (R0) checks that greedy
decoding reproduces the original choice. Only R1 moves the choice. Under R1 the model picks A and
cites the inserted sentence as its reason; under R2 it keeps E and repeats its complaint that
A's profile says nothing of the director's death; under R3, R4 and R0 it keeps E. On this item
the stated reason did work. The forced single-token probability on A, by contrast, barely moves:
0.055 at R0, 0.079 under R1, and 0.089 under R2%
. The two readouts can disagree on a single item, a pattern RQ3 takes up.

Three questions organise the paper. RQ1, content: does supplying the named fact move the choice
more than a length-matched irrelevant sentence at the same profile? This is R1$-$R2 at the named
rival and R3$-$R4 at the unmentioned option. RQ2, location: does the same sentence move the
choice more at the profile the model named than at one it never mentioned? This is R1$-$R3 for
the relevant sentence and R2$-$R4 for the irrelevant one. RQ3, instrument: do the free-text
choice and the probability read agree, and does a judge-free pipeline of string rules measure
what it claims to? In the nearest prior tests a model or a learned editor authors the edit
\citep{atanasova2023faithfulness,chen2024models,mayne2025llms}; here the experimenter does. The
headline numbers come from one run in three designs: Part A, three open 7--8B models at four options;
Part B, three further checkpoints at four options; Part C, Part A's roster at six options. The
answers are mixed, and Table~\ref{tab:summary} states them in one place.

\begin{table}[b]
  \footnotesize
  \caption{What the evidence supports, by research question. Holm $p$ is adjusted over the
  twelve-test discrete family (Table~\ref{tab:contrasts}); a contrast is called supported only
  when it clears that correction. The last column names the check that most qualifies each
  reading. RQ3 has no single contrast to adjust, so its rows report the checks directly.%
  }
  \label{tab:summary}
  \begin{tabular}{p{0.22\linewidth}p{0.12\linewidth}rp{0.50\linewidth}}
    \toprule
    Question & Contrast & Holm $p$ & Reading \\
    \midrule
    RQ1: named fact vs.\ irrelevant control, at the named rival
      & A R1$-$R2 & 0.0210 & Supported; survives every single-model exclusion, and grows when neither sentence names a co-candidate (Section~\ref{sec:audit}). Weakens to uncorrected $p{=}0.099$ on the third of items where repair and control are closest in fluency (Section~\ref{sec:limitations}). \\
      & C R1$-$R2 & 0.1490 & Not supported. \\
    RQ1: the same, at the unmentioned option (the designed contrast)
      & A R3$-$R4 & 0.2428 & Not supported. \\
      & C R3$-$R4 & 0.0167 & Supported, on one model: dropping Mistral takes it to $p{=}0.30$. \\
    RQ1: replication on a second roster
      & B R1$-$R2 & 1.0000 & Not supported. \\
      & B R3$-$R4 & 0.2452 & Not supported; like B R1$-$R2, indistinguishable from an underpowered look at Part A. \\
    \midrule
    RQ2: named fact, named vs.\ unmentioned location
      & C R1$-$R3 & 0.0345 & Supported, but fails once rejections that only rank the rival below the choice are dropped (Section~\ref{sec:audit}), and the probability read points the other way. \\
    RQ2: irrelevant sentence, named vs.\ unmentioned location
      & C R2$-$R4 & 0.0008 & Supported; the strongest result, a location effect with no content claim. The rival's higher starting probability does not explain it in the one testable model, on nine discordant pairs; the tighter band is not significant (Section~\ref{sec:naming-adds}). \\
    \midrule
    RQ3: do the two readouts agree?
      & & & Same sign on all six content cells; opposite on four of six location cells. Three explanations tested, none supported. \\
    RQ3: does the pipeline measure what it claims?
      & & & Validation caught eight defects: three corrected in every number, five quantified, three of those bounded (Section~\ref{sec:audit}). Uncorrected, the parser would report six surviving contrasts, not four. \\
    \bottomrule
  \end{tabular}
\end{table}

Section~\ref{sec:relwork} places the design among faithfulness tests. Section~\ref{sec:method}
describes the corpus, the elicited rejection, the five conditions, the integrity gates, and the
three runs. Section~\ref{sec:results} answers RQ1 and RQ2, including two checks on what the
location effect and the date repairs could mean. Section~\ref{sec:measurement} takes up the
disagreement between the two readouts. Section~\ref{sec:audit} reports how the pipeline was
validated and what validation caught.
Section~\ref{sec:limitations} lists what the design does not yet isolate, and
Section~\ref{sec:conclusion} closes.

\section{Related work}
\label{sec:relwork}

Contrastive explanation, why this and not that, is how people explain choices, structured around
a fact and a foil \citep{miller2019explanation,miller2021contrastive}, with NLP-side formulations
\citep{jacovi2021contrastive,yin2022interpreting}. ERASER's faithfulness is
automatic, from re-running the model on erased input; only plausibility uses human ratings
\citep{deyoung2020eraser}. A contrastive rejection specifies a falsification condition in the
text it produces, reducing that distinction to a string check and a re-ask.

The closest instrument is Atanasova et al.'s counterfactual test of natural language
explanations \citep{atanasova2023faithfulness}: a learned editor inserts a reason that
changes the model's prediction, and faithfulness is read from whether the explanation reflects
that insertion rather than from whether the choice moves. Here the experimenter authors the edit
instead, real corpus text with subject substituted, and reads the model's subsequent choice, with
a length-matched content control (a known distractor \citep{shi2023distracted}) and a location
control unmatched elsewhere. Wiegreffe et al.\ instead correlate a model's label and rationale
sensitivity to one perturbation \citep{wiegreffe2021measuring}. Ding et al.\ likewise treat an explanation as a falsifiable
hypothesis, generating counterfactual edits for vision-language models by altering a cited
concept rather than supplying a missing one \citep{ding2025explanation}. MiCE searches for the
minimal edit that flips a classifier to a specified contrast label \citep{ross2021mice}; here the
model states the edit and the experimenter performs it.

Five further lines test a different mechanism. Turpin et al.\ show chain-of-thought
reasoning can omit an influence that steered the answer \citep{turpin2023language}; Lanham et
al.\ corrupt that reasoning text and check whether the answer changes
\citep{lanham2023measuring}. Chen et al.\ have a separate LLM author the edit and humans predict
the outcome, finding low precision \citep{chen2024models}, though a later study reports otherwise
\citep{mayne2026positive}. Mayne et al.\ have the model author its edit and read its size
\citep{mayne2025llms}. Matton et al.\ estimate concept-level causal effects with a Bayesian
hierarchical model over LLM-authored counterfactuals \citep{matton2025walkthetalk}. None
supplies a fact the model itself named as missing, and in none does the experimenter author the edit
from corpus text.

Self-explanation faithfulness varies by explanation type, model, and task
\citep{madsen2024self}; contrastive rejection is another untested case. Explainers disagree on
the same model \citep{krishna2022disagreement}; our disagreement is within one explainer.
FaithLM contradicts a stated explanation and rechecks the prediction \citep{chuang2026faithlm};
supplying a fact the explanation calls missing is the opposite edit. ALCE scores citations by whether the passages cited entail the statement
\citep{gao2023alce}, and ContextCite traces which context a generation used
\citep{cohenwang2024contextcite}; neither edits text to test a stated claim.

\section{Method}
\label{sec:method}

\subsection{Corpus and items}

Items come from 2WikiMultihopQA \citep{ho2020constructing}. Each item supplies several candidate
entities with short, type-matched profiles: four options in Parts A and B, six in Part C. An
attribute missing from one profile is often present, as a real sentence, in a sibling
profile in the same item, letting every repair come from the corpus itself, for the attributes it
states often enough to source this way: date of birth, date of death, parentage, and director
credits.

\subsection{Elicitation and the properties of elicited rejections}

A model is shown the question and the profiles, picks one, and is asked which candidate it ruled
out and what is missing from that candidate's profile. Asking matters: pooled over three models and the same 40 items, a
contrastive rejection appears unprompted in 20.8\% of items against 72.5\% when asked directly%
, while accuracy stays flat, 68.4\% spontaneous against 68.2\% elicited%
. Asking manufactures the utterance without manufacturing the error, so this paper uses the
elicited form throughout. That pilot had set a threshold of 15 usable items per model, and no
cell reached it: the best reached 12 of 40%
.

About 32\% of specific rejections name a defect the profile carries, 31.6\% spontaneous and
31.8\% elicited%
, a complaint the model's own input refutes. In 18.8\% of attempted items
the rejected candidate is the dataset's correct answer%
, so we exclude these from every repair arm: repairing the gold answer's profile is out of
scope.

\subsection{The repair edit and the five conditions}

When a rejection names an attribute the profile lacks, the design inserts a real
sentence carrying that attribute, drawn from a sibling profile with its subject substituted,
into the target profile. The repair need not be true of the target entity: the claim under test
is only that the profile omits the named attribute, and any sentence stating it falsifies
that claim. Substitution is deterministic: the subject becomes the target entity's name where
that is grammatical, a pronoun where it is not, and sentences where neither works are dropped and
counted; that loss compounds across the repair sentence and the independently chosen control,
14.7\% and 27.2\% of all drops. %

\begin{figure}
  \centering
  \includegraphics[width=0.80\linewidth]{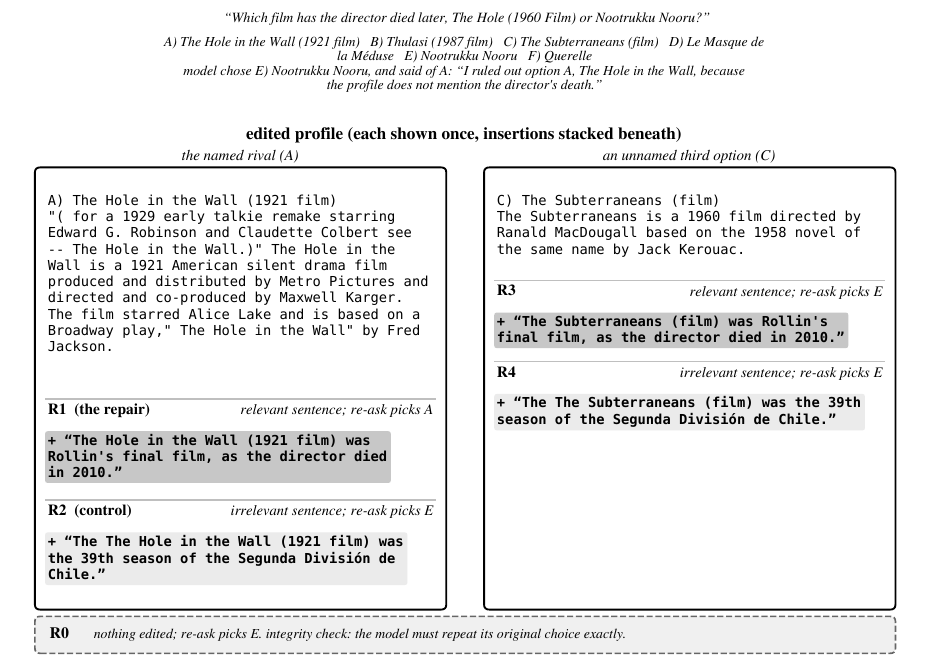}
  \caption{The five conditions as edited content against edited location, on the real built
  six-option item walked through in Section~\ref{sec:intro}; the banner lists all six candidates.
  Columns are locations, the named rival and the third option; each
  profile shows its two insertions stacked beneath it (R1/R3 relevant, R2/R4 irrelevant
  length-matched control), with the option the re-ask then picks. R0, below, is
  the unedited integrity check. Profile text is verbatim from the corpus,
  punctuation artifacts included; inserted sentences are corpus sentences with the subject
  substituted; the doubled article in R2 and R4 is one such substitution artifact, found in 3 of
  Part C's 563 controls.%
  }
  \label{fig:design}
\end{figure}

Five conditions apply this insertion to different targets and content. R0 re-asks with
nothing edited, an integrity check: greedy decoding means the choice must reproduce exactly.
Figure~\ref{fig:design} shows the remaining four conditions, R1 through R4, as a 2$\times$2 of
content against location on the item of Section~\ref{sec:intro}, with R2 and R4 the load-bearing controls. If
either moves the choice as often as its relevant counterpart, editing a paragraph disturbs the
model regardless of what it says. R1$-$R2 and
R3$-$R4 isolate the effect of content at fixed location (RQ1); R1$-$R3 and R2$-$R4 isolate the effect
of location at fixed content (RQ2). Balkir et al.\ define necessity and sufficiency over removing a
feature \citep{balkir2022necessity}; these conditions add one instead, to a profile that already
lacks it, so neither label transfers unchanged. The primary measure
throughout is $P(\text{the model now chooses the edited option})$ against R0, paired per item.

R1 and R3 draw that sentence only from sibling options inside the same item (pool median
1--2). R2 and R4 draw from those siblings plus the entire corpus index, ordered by token-length
proximity with no preference for a sibling (pool median 1{,}430 to 7{,}241 by part). So
in 1{,}263 of 1{,}264 built items the control sentence comes from outside the item. Read by
whole-word match on the 1{,}070 items whose sentences rebuild offline, it states a parentage
relation in 20--30\% of items per part, against 3--4\% for the repair. %

A switch under a controlled input change is output consistency
\citep{parcalabescu2024selfconsistency}, compatible with mechanisms other than sensitivity to the
criterion (Section~\ref{sec:limitations}). Work on knowledge conflicts between supplied context
and parametric memory bears on this \citep{longpre2021entity,xie2024adaptive}; a substituted
sentence checked against the entity's attested true value has
never populated on this corpus, zero of 295, 54 and 125 built items across three option counts,
leaving open whether parametric belief enters here%
.

\subsection{Integrity gates}

Eight deterministic checks run before any generation, since a failure means the edited
passages are wrong; any single failure excludes that R1/R2/R3/R4 combination, and exhausting every
combination drops the item. Table~\ref{tab:gates} enumerates all eight.
Gate 8 is the largest single source of exclusion in every part of this design, its
selection consequences discussed in Section~\ref{sec:limitations}. %
An early version picked that option blindly and failed on the 65.1\% of items already correct
at stage one%
; the reported runs instead prefer, among eligible candidates, whichever already lacks the named
attribute, recovering 25--35\% more built items per model%
.

\begin{table}
  \footnotesize
  \caption{The eight deterministic integrity gates, from the implementation.
  }
  \label{tab:gates}
  \begin{tabular}{cp{0.85\linewidth}}
    \toprule
    \# & Predicate (conditions it applies to) \\
    \midrule
    1 & Named attribute absent from the rival's profile before R1's edit, present after (R1) \\
    2 & R2 edits the rival's profile only, without introducing the named attribute (R2) \\
    3 & R1 and R2's inserted sentences are within 20\% of each other in word count (R1, R2) \\
    4 & No inserted date leaves a birth date on or after a death date, in any condition (R1--R4) \\
    5 & The inserted sentence contains no other candidate's full title verbatim, in any condition (R1--R4) \\
    6 & The gold answer's profile is unchanged in every condition (R0--R4) \\
    7 & Option order, titles, and question text are identical across all five conditions (R0--R4) \\
    8 & R4 edits the same option as R3, without introducing the named attribute (R3, R4) \\
    \bottomrule
  \end{tabular}
\end{table}

Two of these gates test less than their names suggest;
Section~\ref{sec:audit} quantifies why for both.

\subsection{The probability measure and a corrected probe}

Alongside the discrete choice, a forced single-token re-ask reads the model's probability on
each candidate letter, giving a continuous $\Delta p$ per item, exploratory only. The probe
places its instruction in the existing final user turn. A first form added a separate user
turn, which some chat templates silently accept and one rejects outright; validation caught
this during the largest run, and the part already generated was regenerated before any later
part began. The form is not cosmetic: on the one model where both were kept, 178 to 207 of 504
non-R0 rows differ by an order of magnitude on the edited-option probability%
. No row from the first form enters any figure this paper reports%
.

\subsection{Three runs}

A first, small run found the choice moving toward a third option at a rate close to the direct
repair's%
, too small for significance, motivating the matched irrelevant-content control (R4) and
this 2$\times$2 design. A second run built that design at scale ($n{=}295$) but left location
open, and its rival's-profile effect was substantially one model's: Qwen supplied 77.8\% of the
pooled net swing from 32\% of the items, taking the contrast to $p{=}0.54$ on exclusion%
. The third and largest run supplies the headline numbers, with the gate-eight selection
preference and the corrected probe above, across three designs: the original
three-model roster (Qwen2.5-7B-Instruct, Llama-3.1-8B-Instruct,
Mistral-7B-Instruct-v0.3) at four options with the corrected probe (Part A, $n{=}387$ of 1204),
three checkpoints untried before, again at four
options (Qwen2.5-14B-Instruct-AWQ, Qwen2.5-3B-Instruct, and DeepSeek-R1-Distill-Qwen-14B-AWQ;
Part B, $n{=}314$ of 875), and the original roster at six options (Part C, $n{=}563$
of 1293)%
. R0 was clean on all nine model-part cells, 0 flips across 1264 rows%
, so generation is deterministic within a run at this scale.

\subsection{Code and data release}
\label{sec:release}

Code and per-item records for all three parts are released at
\url{https://github.com/ArchitRastogi20/contrastive-rejection-test} under the MIT licence. The release
carries the experiment harness, deterministic scoring rules, analysis scripts, and one
JSON record per item per condition including every raw model response. Every reported number in
Sections~\ref{sec:results} to~\ref{sec:audit} regenerates from those records via the seeded
modules under \texttt{harness/}, mapped number by number in the release README, so a rerun
reproduces the intervals exactly; the audits that rebuild inserted sentences
(Sections~\ref{sec:results-dates} and~\ref{sec:audit}) also need a local copy of the
2WikiMultihopQA validation split.%

\section{Results}
\label{sec:results}

The population under test throughout this section is items surviving all eight gates, drawn only
from elicited rejections, a subpopulation skewed toward stage-one-correct items in every
part (Section~\ref{sec:limitations} quantifies the skew). Each item's R0 response is paired
against its response under one edited condition, read as a choice flip or a probability change.
Every interval
is a percentile bootstrap over item-level paired differences, 10{,}000 resamples at seed
20260822, pooled across models with no cluster term; discrete tests are exact McNemar with a
matched-pairs odds ratio and a log-scale Wald interval. Resampling
item ids instead of rows as clusters leaves every content-contrast interval unchanged to two
decimals.%

Built-item yield rose from 32.1\% of attempted items at four options to 43.5\% at six%
, consistent with gate eight's logic: more candidates make a third option already lacking the
named attribute easier to find. Figure~\ref{fig:forest} and Table~\ref{tab:contrasts} give
every number behind Table~\ref{tab:summary}.

\subsection{RQ1: the content effect at each location}
\label{sec:results-necessity}

\begin{figure}
  \centering
  \includegraphics[width=0.86\linewidth]{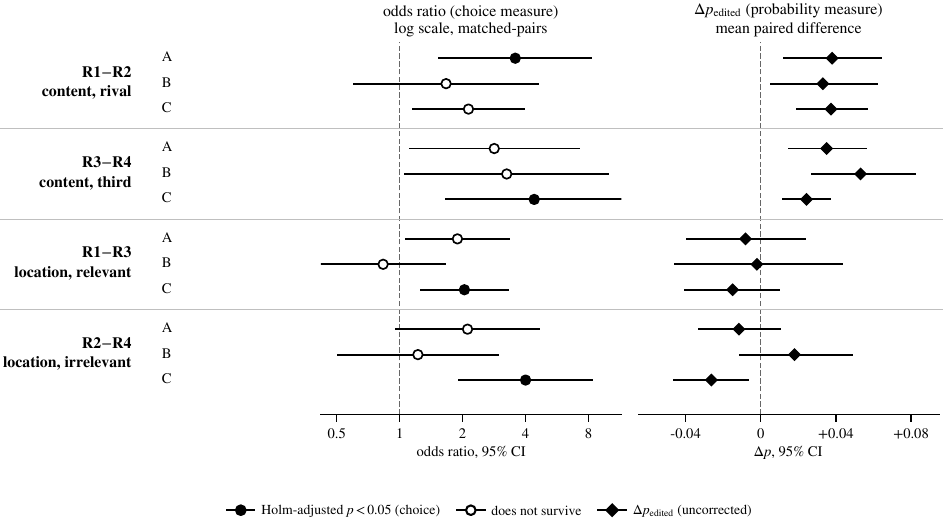}
  \caption{Matched-pairs odds ratios (choice measure, log scale, left) and mean paired
  $\Delta p_{\text{edited}}$ (probability measure, right), pooled across models, for all four
  contrasts in all three parts. Solid circles survive Holm correction over the twelve-test
  discrete family; open circles do not. Diamonds are the continuous measure, outside the Holm
  family throughout (probe coverage per model is given in Section~\ref{sec:measurement}).
  Table~\ref{tab:contrasts} gives the numbers behind every marker.%
  }
  \label{fig:forest}
\end{figure}

Figure~\ref{fig:forest} shows all four contrasts pooled across Parts A, B, and C, on both
measures. Four of the twelve
discrete contrasts clear Holm correction: A
R1$-$R2, content at the profile the model actually named; C R3$-$R4, the same content at the
third option already lacking the attribute; C R2$-$R4, matched
irrelevant content compared across those same two locations; and C R1$-$R3, relevant content
compared across the named and unnamed locations%
. A R1$-$R2 is the most defensible of the four: it survives every single-model exclusion, exact
$p$ 0.0118 dropping Llama-3.1-8B-Instruct, 0.0072 dropping Mistral-7B-Instruct-v0.3, 0.0347
dropping Qwen2.5-7B-Instruct%
. C R2$-$R4 survives every exclusion the same way, exact $p$ 0.0118, 0.0012, 0.0002 dropping each
model in turn. R2 and R4 carry the
identical irrelevant sentence, so this measures where an edit lands rather than what it says%
. C R3$-$R4 clears Holm ($p{=}0.0167$) but rests on one model: dropping
Mistral-7B-Instruct-v0.3 alone takes it to $p{=}0.30$%
.

\begin{table}
  \footnotesize
  \caption{Matched-pairs contrasts behind Figure~\ref{fig:forest}, all four comparisons in all
  three parts. OR is the matched-pairs odds ratio $b/c$
  with an unadjusted Wald interval on the log scale; Holm $p$ is adjusted over the twelve-test
  discrete family. The two measures do not share a denominator: the discrete read needs a
  parseable answer (paired $n$ 377--381 in A, 309--310 in B, 547--551 in C) and the continuous
  read needs a complete letter probe (paired $n$ 299--305 in A, 229 in B, 403--412 in C).
  DeepSeek-R1-Distill-Qwen-14B-AWQ contributes to every
  discrete number in Part B and none of the continuous ones, since its probe never completes.
  $\Delta p$ is reported at 95\% and, like the figure's diamonds, sits outside the Holm family.}
  \label{tab:contrasts}
  \begin{tabular}{llrrlrl}
    \toprule
    Part & Contrast & $b$ & $c$ & OR [95\% CI] & Holm $p$ & $\Delta p$ [95\% CI] \\
    \midrule
    A & R1$-$R2 & 25 & 7  & 3.57 [1.54, 8.26]  & \textbf{0.0210} & $+0.0380$ [$+0.0121$, $+0.0640$] \\
      & R3$-$R4 & 17 & 6  & 2.83 [1.12, 7.19]  & 0.2428 & $+0.0350$ [$+0.0149$, $+0.0560$] \\
      & R1$-$R3 & 34 & 18 & 1.89 [1.07, 3.34]  & 0.2428 & $-0.0081$ [$-0.0394$, $+0.0236$] \\
      & R2$-$R4 & 19 & 9  & 2.11 [0.96, 4.67]  & 0.3486 & $-0.0116$ [$-0.0331$, $+0.0103$] \\
    \cmidrule(lr){1-7}
    B & R1$-$R2 & 10 & 6  & 1.67 [0.61, 4.59]  & 1.0000 & $+0.0331$ [$+0.0054$, $+0.0621$] \\
      & R3$-$R4 & 13 & 4  & 3.25 [1.06, 9.97]  & 0.2452 & $+0.0531$ [$+0.0271$, $+0.0821$] \\
      & R1$-$R3 & 15 & 18 & 0.83 [0.42, 1.65]  & 1.0000 & $-0.0020$ [$-0.0454$, $+0.0431$] \\
      & R2$-$R4 & 11 & 9  & 1.22 [0.51, 2.95]  & 1.0000 & $+0.0180$ [$-0.0113$, $+0.0484$] \\
    \cmidrule(lr){1-7}
    C & R1$-$R2 & 32 & 15 & 2.13 [1.16, 3.94]  & 0.1490 & $+0.0374$ [$+0.0193$, $+0.0565$] \\
      & R3$-$R4 & 22 & 5  & 4.40 [1.67, 11.62] & \textbf{0.0167} & $+0.0244$ [$+0.0120$, $+0.0370$] \\
      & R1$-$R3 & 51 & 25 & 2.04 [1.26, 3.29]  & \textbf{0.0345} & $-0.0150$ [$-0.0402$, $+0.0095$] \\
      & R2$-$R4 & 36 & 9  & 4.00 [1.93, 8.30]  & \textbf{0.0008} & $-0.0263$ [$-0.0463$, $-0.0070$] \\
    \bottomrule
  \end{tabular}
\end{table}

The contrast this design was built to detect is Part A's discrete R3$-$R4 contrast, the content
effect at the third option. It does
not clear Holm correction, $p{=}0.2428$%
. Dropping Qwen2.5-7B-Instruct makes the contrast
stronger, exact $p$ 0.0347 to 0.0063; dropping Llama-3.1-8B-Instruct ($p{=}0.1435$) or
Mistral-7B-Instruct-v0.3 ($p{=}0.3323$) is what removes it%
.

No run of this design establishes a content effect at both locations at once, in either part.

No pooled contrast shows significant between-model heterogeneity: Cochran's $Q{=}0.81$
($I^2{=}0\%$) for A R1$-$R2, 0.78 (0\%) for C R2$-$R4, 3.02 (33.9\%) for A R3$-$R4, 2.84 (29.6\%)
for C R3$-$R4, all $p{>}0.2$%
. With three models, $Q$ has little power, so this is weak evidence of homogeneity.

\subsection{RQ1 on a second roster: no discrete replication}
\label{sec:results-partb}

Part B tested three checkpoints never run before, at four options, to test whether the content
effects generalise. Neither survives Holm correction on Part B's discrete measure
(Table~\ref{tab:contrasts}), though R3$-$R4's own interval excludes an odds ratio of one, 3.25
[1.06, 9.97]: at $n{=}310$ a twelve-test correction stops it rather than the point estimate%
.
Resampling each part within itself, since a pooled resample would treat a between-roster
difference as noise, both roster differences include zero: R1$-$R2's is $+0.035$
[$-0.004$, $+0.073$], R3$-$R4's is $-0.000$ [$-0.036$, $+0.035$]%
. Neither contrast's non-replication can be distinguished from an underpowered look at Part A's
own effect. The probability measure, which cannot read the reasoning model
(Section~\ref{sec:measurement}), moves in the content effect's direction on the other two
(Figure~\ref{fig:forest})%
. All three Part B checkpoints are Qwen lineage, the family the second run's analysis
flagged as carrying most of the rival's-profile effect: two Qwen2.5 sizes and a reasoning model
distilled onto a Qwen architecture%
. So no independent lineage beyond Llama and Mistral was tested, and the
six-model scope of this paper is, on independent architectures, closer to three.
A check on two more independent lineages, Phi-3.5-mini-instruct (3.8B) and
Granite-3.0-8B-Instruct, pooled at 186 items, repeats both content contrasts on the continuous
measure: R1$-$R2 $+0.0517$ [0.0185, 0.0866], R3$-$R4 $+0.1094$ [0.0674, 0.1557]%
. Its discrete measure holds direction on R1$-$R2 without significance ($p{=}0.096$) and is
significant the other way on R1$-$R3 ($p{=}0.001$), opposite its own continuous reading%
. This repeats the content effect on the continuous measure only; the two-measure disagreement
recurs.

\subsection{RQ2: the location contrasts}
\label{sec:naming-adds}

R1$-$R3 asks whether relevant content moves the choice more at the location a model named than
at the one it did not; R2$-$R4 asks the same question of the irrelevant control. Part C's R1$-$R3
clears Holm correction, $p{=}0.0345$; Part A's stays non-significant at $0.2428$%
. The continuous measure disagrees: Part C's R1$-$R3 $\Delta p$ is
$-0.0150$ [$-0.0402$, $+0.0095$], opposite in sign and including zero (Section~\ref{sec:measurement-disagree}).

C R2$-$R4 is the opposite case: content is held fixed and only location varies, yet it clears
Holm correction most comfortably ($p{=}0.0008$, Section~\ref{sec:results-necessity}) and
survives every single-model exclusion. R1$-$R2 and
R3$-$R4 hold location fixed, so a location artefact cannot manufacture either result. What it does
undercut is the comparison between them, since an edit's effect here depends on where it lands.

Every re-ask is a fresh single-turn prompt, so the model never sees its earlier rejection; the
rival is simply the option the model treats as the live alternative. On one reading, the
runner-up is where any disturbance is felt first, whatever its probability. On the other, the
rival starts higher, at about twice the third option's unedited probability
(Table~\ref{tab:baseline}), and needs a smaller push to cross the argmax threshold. The two may
be one mechanism; the second is testable by restricting to items where the locations start close. In the band
where the two R0 probabilities lie within a factor of three of each other and both exceed 0.01,
Part C holds 87 items, and the contrast keeps its direction and its nominal, uncorrected
significance, $b{=}8$ against $c{=}1$, exact $p{=}0.039$, risk difference $+0.080$ [$+0.023$,
$+0.149$]; Part A's band holds 74 items, $b{=}9$ against $c{=}2$, $p{=}0.065$. In the tighter band,
within a factor of 1.5, neither part is significant, $b{=}4$ against $c{=}0$ in C and 7 against 1
in A%
. The bands are almost entirely one model's: Llama-3.1-8B-Instruct supplies 85 of Part C's 87
items and 72 of Part A's 74, because the other two models put near-zero probability on one of
the two locations in nearly every item%
. So this is a within-Llama check on nine discordant pairs, against the roughly 728 an 80\%
power calculation would need%
, and it does not reach Qwen or Mistral, where the pooled effect also lives. It leaves the
baseline reading not supported and not ruled out. On this contrast the probability measure
points the other way, its interval entirely below zero (Section~\ref{sec:measurement}).

\subsection{Do date repairs change the right answer?}
\label{sec:results-dates}

A date inserted into a profile can change which option the question itself favours, so a
switch toward the repaired option might be appropriate use of new evidence rather than
sensitivity to the stated reason. Two checks bound this. First, the content contrasts were
recomputed inside and outside the stratum of order questions with a date repair, 594 built
items. In Part A both are larger outside it, R1$-$R2 OR 5.33 against 2.25 and R3$-$R4 5.00
against 1.75; in Part C R3$-$R4 is 19.0 against 2.60, while R1$-$R2 is the one reversal, 2.00
against 2.17, with 38 of its 47 discordant pairs inside the stratum%
. The effects are not confined to items where the date could bear on the answer. Second, for
the 1{,}070 built items that can be rebuilt offline, each inserted year was compared with the
competitor's year in the direction the question asks. Most rows cannot be scored: 352 questions
compare a related entity's date (405 of the stratum's 594 items ask which film's director died
or was born first, and a death date on a film's profile may be read as the director's), 82 name
only one option, and the edited option is absent from the question in 20 rows under R1 and 52
under R3. Of the 37 scorable R1 rows, 12 tie, because the only sibling stating the date is the
competitor itself; 6 become correct and one of those switched; 18 stay incorrect and one
switched; one is unreadable. So of 59 R1 switches inside the stratum, 4 fall on scorable rows and one lands on an
option the inserted date makes correct; of 36 R3 switches, 2 are scorable and neither does%
. This bounds question relevance only on the items that name two people directly. For the
director questions, most of the stratum, it stays open, and only a gate that tests the inserted
value against the question, not against the profile, would close it.

\section{Measurement}
\label{sec:measurement}

\subsection{Two measures and three tested explanations}
\label{sec:measurement-disagree}

A first-token probability read and a generated-text answer are known to disagree, mismatching
above 60\% on one subjective dataset \citep{wang2024firsttoken}, with the text
answer the more robust under perturbation \citep{wang2024firsttoken,wang2024lookatthetext}.
The two measures agree in sign on all six
content-contrast cells and diverge on four of six location cells%
; the reversal on the matched R2$-$R4 contrast does not follow from a first-token artefact
(Explanation 2 below). A binary flip read measures strictly less than a distribution read
\citep{siegel2024probabilities}, and continuous metrics, though more diagnostic, are more
model-sensitive \citep{zaman2025causal}.

Before any edit the rival carries 1.94$\times$ the third option's probability in Part A and
2.36$\times$ in Part C, and 1.44$\times$ in Part B, whose probabilities run far lower overall
(Table~\ref{tab:baseline})%
. The gap is expected: the model names as rejected the option it ranks second. Probability
mass over lettered options is independently shaped by position and by a prior over the answer
symbol \citep{zheng2024robust,pezeshkpour2024order}, which an argmax inherits. Each item's option
order is fixed by a per-item seeded shuffle, reused unchanged by every condition, so letter
assignment cannot vary within a pair. %
That bias is large where it is free to act: pooled R0 choice rejects a uniform distribution over
letters in both Part A and Part C ($\chi^2$, $p<0.001$ in both), and one model,
Llama-3.1-8B-Instruct, drives most of the pooled deviation, within-model $\chi^2$ 129.9 on Part A%
.

\begin{table}
  \small
  \caption{R0 (unedited) probability by target location, pooled across models. Part B excludes
  the reasoning model, whose probe never completes. Continuous results throughout this paper are
  complete-case, and probe coverage differs from the choice measure's coverage
  (Section~\ref{sec:probe-portability}).
  }
  \label{tab:baseline}
  \begin{tabular}{lrr}
    \toprule
    Part & rival mean & third-option mean \\
    \midrule
    A & 0.145 & 0.075 \\
    B (reasoning model excluded) & 0.046 & 0.032 \\
    C & 0.140 & 0.060 \\
    \bottomrule
  \end{tabular}
\end{table}

\textbf{Explanation 1: a ceiling effect.} An item where the rival already carries high
probability needs only a small push to cross the discrete threshold but has little headroom
left, so the two measures could diverge without a real effect difference. Normalising by the
available headroom should then shrink the asymmetry; instead it grows. Part C's pooled R2$-$R4
gap, a continuous quantity no choice-parsing rule touches, moves from $-0.026$, 95\% CI $[-0.046,-0.007]$ raw to
$-0.082$, 95\% CI $[-0.139,-0.036]$ once trimmed and renormalised this way%
, the opposite of what a ceiling predicts, and none of the battery's three tests supports it.

\textbf{Explanation 2: the discrete measure reads the wrong quantity.} An argmax discards
within-ranking movement, governed by margin against the best competitor rather than raw
probability, so reading margin might reconcile the two measures. It does not: Part C's pooled
R2$-$R4 margin sides with the probability read and against the flip counts, $\Delta$margin $-0.0400$ [$-0.0783$, $-0.0031$]%
.

\textbf{Explanation 3: a few large moves outweigh many small threshold crossings.} McNemar's
count is a majority vote over sign changes; a mean can be dominated by a handful of large moves
in the opposite direction. In Part C, though, the sign count opposes the flip direction,
181 items moving positive against 226 negative, against the mean%
. The leave-one-out check in Section~\ref{sec:results-necessity} locates the disagreement
instead: C R3$-$R4's discrete result depends on Mistral-7B-Instruct-v0.3, dropping it alone
takes $p$ to 0.30, while its continuous result, and A R3$-$R4's and A R1$-$R2's, depend on
Llama-3.1-8B-Instruct instead%
. The two measures are not reading the same subset of the roster.

\subsection{Probe portability across architectures}
\label{sec:probe-portability}

The probability measure depends on a forced single-token decode finding every candidate's letter
in the top of the model's output distribution, itself presuming the model binds answer symbols
to option content \citep{robinson2023leveraging,xue2024strengthened}. That decode is complete on every row for four of
the six models tested, but on only 55.4\% of one model's rows at four options and 32.8\% at six%
, and on zero of 425 rows for a reasoning model%
. The reasoning model's failure is structural: it always opens its reply with a
reasoning block \citep{deepseekai2025r1}, so the single decoded token is always the start of that
block, never a letter. The other model's failure appears tied to vocabulary tokenization and
candidate count rather than the probe's construction, since the same model's discrete-choice
read stays complete throughout. Completion also varies by condition for the
degraded model, in Part A and Part C only, reproducible by
\texttt{harness/audit\_probe\_missingness.py} in the release%
. The continuous results are therefore complete-case under a missingness pattern that is not
random for that model, and only Part C's R2$-$R4 has been checked for the resulting bias.

\section{Validating the instrument}
\label{sec:audit}

Every measurement here is a deterministic string rule: the parser that reads a choice, the cues
that decide whether a profile states an attribute, the gates that accept an edit. Such rules fail
silently, so each was validated against the records it reads: the parser against an independent
ground-truth rule, the gates by rebuilding every inserted sentence from the corpus, each cue by
reading what it matched. Validation caught eight defects. Three are corrected in every reported
number; five need new generation to correct, so they are quantified, and three are bounded.

The three corrected defects are in the choice parser. It returned a rejected option as chosen
in 17.1\% of adjudicable responses; its letter range stopped at four options while Part C offers
six, which left 134 of 155 unreadable Part C rows readable once widened; and it lost a chosen
option's trailing parenthetical disambiguator on restatement, 21 of 2{,}960 adjudicable
stage-three rows, 16.2\% of the 130 exposed%
. The worked example of Section~\ref{sec:intro} shows the first: its R2 response opens ``E)
Nootrukku Nooru'' and goes on to rule out A, and the uncorrected rule read that as a choice of
A. The corrected parser has zero disagreements against the ground-truth rule over 10{,}038 rows
and zero regressions, with every test passing%
. On the uncorrected parser's output the same Holm ladder reports six surviving contrasts
instead of four: which conclusions a faithfulness study reaches can turn on a string rule nobody
thought to test.

The five that remain are these. Gate three's word-count check (Table~\ref{tab:gates}) cannot
fail on a built item by construction, since \texttt{check\_integrity} tests the metric the
search minimises, confirmed on 1{,}070 rebuilt items; a real-tokenizer check is not possible
offline. Gate five's verbatim-title test misses a co-candidate mention at rates tracking both
contrasts' treatment axis, 28.0\% against 4.3\%, since the relevant sentence comes from a
sibling profile and the irrelevant one does not%
. The cue that classifies a named attribute as parentage matches by unbounded substring, on
``son'' inside ``person'' or ``Johnson'', misclassifying 112 of 127 parentage-attribute items%
, so a stratum selected by that label is not the relation-matched case it appears to be. In
15.7\% of built items the rejection only ranks the rival below the choice rather than asserting
absence, and dropping them flips C R1$-$R3 from clearing Holm correction to failing it%
. The attribute-presence rule matches its cue as an unbounded substring too, so ``studied''
followed by a year passes as a date of death, in 6 of the 648 date-repair sentences rebuilt
under each of R1 and R3 (0.9\%)%
. Two further faults, the probe's first form and blind gate-eight selection
(Section~\ref{sec:method}), were caught before the reported runs, so no reported number carries
them, and they are not counted among the eight.

Three of the five bear directly on the content contrasts, and each can be bounded by recomputing
R1$-$R2 and R3$-$R4 on the rebuilt items where the two sentences of a pair agree on the confound,
with \texttt{harness/audit\_confound\_bounds.py}. Where neither sentence names a co-candidate,
all four Part A and Part C content effects grow against their rebuilt-item baseline, A R1$-$R2
from OR 3.14 to 4.75 ($b{=}19$, $c{=}4$) and C R3$-$R4 from 4.40 to 9.00: the confound that
tracks the treatment most closely works against the effect. Where repair and control agree on
stating a parentage relation, both Part C effects grow and both Part A effects weaken without
reversing, R1$-$R2 to OR 2.33 ($p{=}0.12$). Dropping the six embedded-cue date repairs moves no
count by more than one. On the intersection of all three, all four keep an odds ratio above 4 at
uncorrected $p \le 0.04$; Part B's, on at most twelve discordant pairs, move both ways%
. These strata are post hoc and uncorrected, and do not match on fluency.

\section{Limitations}
\label{sec:limitations}

This work spans one corpus and one language; four relation types, dates of birth and death,
parentage, and director credits, account for 98\% of built items%
. Every rejection tested here was elicited (Section~\ref{sec:method}), and the six models reduce
to about three independent lineages (Section~\ref{sec:results-partb}).

A switch after repair admits several mechanisms besides sensitivity to the stated criterion,
among them changed salience, inconsistency-induced uncertainty, an edit-detection cue, or a
generic preference for relation-bearing text. People confabulate similarly: handed back a face
they had not chosen, at most 26\% noticed the swap; the rest justified a choice they never
made \citep{johansson2005failure}%
, and a re-ask cannot rule this out, being itself a second decision. An edit also moves the input off-distribution \citep{hase2021ood,feng2018pathologies}, here by
about 1.2 nats/token in every model tested (pooled R1$-$R2
$-1.114$ [$-1.182,-1.045$], $n{=}925$; R3$-$R4 $-1.181$ [$-1.246,-1.115$], $n{=}931$)%
. In the closest-fluency tercile R1$-$R2 weakens to OR 2.00, $p{=}0.0987$, while R3$-$R4 stays
nominally significant, OR 3.00, $p{=}0.0414$, both uncorrected%
.

The repair sentence and its control therefore differ on four axes: the named attribute itself,
by design; fluency; whether the sentence names a third candidate; and whether it states a
parentage relation. Post-hoc matching on the last two preserves the direction of every Part A
and Part C content effect (Section~\ref{sec:audit}); matching on fluency weakens the Part A named-rival effect. Until repair and control
are matched on fluency within one experiment, R1$-$R2 and R3$-$R4 bound an effect rather than
establish one.

One comparison is impossible in any run of this design: gate eight tests the post-edit R4
profile for the named attribute, so an option that already carries it can never be a target.
A variant with random target selection, run on one model, was underpowered and did not exclude
zero%
. Yet that forbidden case, an option already stating the named fact before any edit, is what
would separate a runner-up effect from a general location effect: editing elsewhere, a broader
criterion predicts movement toward that option and a location effect predicts none. Building it
means replacing gate eight, under one GPU hour at this study's measured throughput%
; no run reported here does.

Gate eight also skews selection: built items with a readable stage-one answer were correct in 72.9\%
against 50.2\% of the attempted pool in Part A, 85.6\% against 67.6\% in B, 48.0\% against
42.0\% in C%
.

\section{Conclusion}
\label{sec:conclusion}

The instrument, and the validation that bounds it (Section~\ref{sec:audit}), is what this paper
transfers. On RQ1, supplying the named fact moves the choice at the named profile, surviving
every single-model exclusion, the twelve-test correction, and post-hoc matching on co-candidate
mention, though not conditioning on fluency (Section~\ref{sec:limitations}). The same edit at an option no
model mentioned, the contrast the design was built to detect, survives in Part C on one model and
not in Part A. On RQ2, Part C's R1$-$R3 clears correction ($p=0.0345$) but fails once rank-only
rejections are excluded, and the strongest result carries no content claim at all: an edit's
placement alone moves the choice, so every content result needs checking against its location's
disruptiveness. On RQ3, the two readouts part ways exactly where location enters, and validating
the pipeline's string rules changed which conclusions survive.

Whether the model's stated reason caused the original decision is still unknown, and so, until the
confounds of Section~\ref{sec:limitations} are separated, is whether this design's own effect is
the one that would show it.

\section*{Declaration on Generative AI}

During the preparation of this work, the author used Claude Sonnet 5 in order to: Paraphrase and
reword, and Improve writing style. After using this tool, the author reviewed and edited the
content as needed and takes full responsibility for the publication's content.

\bibliography{refs}

\end{document}